\pdfoutput=1

\documentclass[11pt]{article}

\usepackage[final]{acl}

\usepackage{times}
\usepackage{latexsym}

\usepackage[T1]{fontenc}

\usepackage[utf8]{inputenc}

\usepackage{microtype}

\usepackage{inconsolata}

\usepackage{graphicx}
\usepackage{booktabs}
\usepackage{multirow}
\usepackage{amssymb} 
\usepackage{caption} 
\usepackage{amsmath}
\usepackage{tcolorbox}
\usepackage{subcaption}

\newcounter{observationcount}
\newtcolorbox{observation}[2][]{%
  colback=blue!5!white,
  colframe=blue!75!black,
  fonttitle=\bfseries,
  before upper={\stepcounter{observationcount}},
  title=Observation \#\theobservationcount\if\relax\detokenize{#2}\relax\else: #2\fi,
  #1
}

\title{Grouped Sequency-arranged Rotation: 
Optimizing Rotation Transformation for Quantization for Free}

\author{Euntae Choi\thanks{these authors contributed equally.}, Sumin Song\footnotemark[1], Woosang Lim\footnotemark[1], Sungjoo Yoo \\
  Seoul National University \\
  \texttt{euntae.choi175@gmail.com, songsm921@snu.ac.kr,} \\ \texttt{ftyg656512@snu.ac.kr, sungjoo.yoo@gmail.com} \\}

\begin{document}
\maketitle
\begin{abstract}
Large Language Models (LLMs) face deployment challenges due to high computational costs, and while Post-Training Quantization (PTQ) offers a solution, existing rotation-based methods struggle at very low bit-widths like 2-bit. We introduce a novel, training-free approach to construct an improved rotation matrix, addressing the limitations of current methods. The key contributions include leveraging the Walsh-Hadamard transform with sequency ordering, which clusters similar frequency components to reduce quantization error compared to standard Hadamard matrices, significantly improving performance. Furthermore, we propose a Grouped Sequency-arranged Rotation (GSR) using block-diagonal matrices with smaller Walsh blocks, effectively isolating outlier impacts and achieving performance comparable to optimization-based methods without requiring any training. Our method demonstrates robust performance on reasoning tasks and Perplexity (PPL) score on WikiText-2. Our method also enhances results even when applied over existing learned rotation techniques.
\end{abstract}

\section{Introduction}
Large Language Models (LLMs), despite their widespread success, face deployment challenges due to high computational costs, particularly in resource-constrained settings. Quantization, which reduces the numerical precision of model parameters, offers a viable solution by decreasing model size and accelerating computation with minimal accuracy loss. Post-Training Quantization (PTQ) is especially attractive as it avoids costly retraining.

Within PTQ for LLMs, rotation-based methods like QuaRot~\cite{quarot} are common but suffer severe performance degradation at low bit-widths, such as 2-bit weight quantization (W2), exhibiting high Perplexity (PPL) of 20.29 on WikiText-2~\cite{wiki2}. Subsequent methods like SpinQuant~\cite{spinquant} (PPL of 16.45) and OSTQuant~\cite{ostquant} (PPL of 10.97) improve accuracy using learnable rotation or scaling matrices, but require additional optimization phases, diminishing the core benefit of PTQ.

To address this, we propose a novel, training-free approach to construct an improved rotation matrix for LLM quantization. Our method leverages the Walsh matrix by rearranging the rows of the Hadamard matrix so that the sequency is sorted in ascending order. This clusters similar frequency components, reducing intra-group variance and quantization error compared to the standard Hadamard matrix used in QuaRot, improving PPL to 15.38.

Furthermore, inspired by local rotation techniques~\cite{duquant, duarot}, we introduce Grouped Sequency-arranged Rotation (GSR). The GSR employs a block-diagonal matrix with smaller Walsh matrices, effectively isolating outlier impacts within each quantization group. This significantly enhances performance, achieving a PPL of 11.59 and an average zero-shot tasks accuracy of 42.44\% – comparable to optimization-based methods without requiring training. Our approach also improves when applied to existing learning-based methods like SpinQuant and OSTQuant.

\section{Preliminaries}
\subsection{Walsh-Hadamard Transform and Sequency}
A Hadamard matrix with a size of a non-negative power of two is usually constructed by Sylvester's method as follows:
\begin{equation}\label{eqn:sylvester}
\textbf{H}_2=\frac{1}{\sqrt{2}}\begin{bmatrix}1 & 1 \\ 1 & -1\end{bmatrix} \quad \text{and} \quad \textbf{H}_{2^n}=\textbf{H}_2 \otimes \textbf{H}_{2^{n-1}}.
\end{equation}
A Walsh matrix is derived by applying the bit-reversal and the Gray-code permutation to the Hadamard matrix~\cite{walsh}.

Sequency is the number of sign flips in a row of such matrices. The Walsh matrix follows sequency ordering where the sign flips of each row are arranged in ascending order. In contrast, the Hadamard matrix is in natural ordering, where the sequency value of the i-th row is defined as follows:
\begin{equation}\label{eqn:naturalordering}
\text{S}(\text{i})=\text{bit\_count}(\text{i}\oplus(\text{i}>>1)).
\end{equation}
For instance, the rows of a Hadamard matrix of size 8 have 0, 7, 3, 4, 1, 6, 2, and 5 sequency values.

Such matrices serve as a transform by themselves, and we call each row (or column) a sequency filter.

\subsection{Rotation for LLM Quantization}
\begin{figure}[] 
    \centering
    \includegraphics[width=\columnwidth]{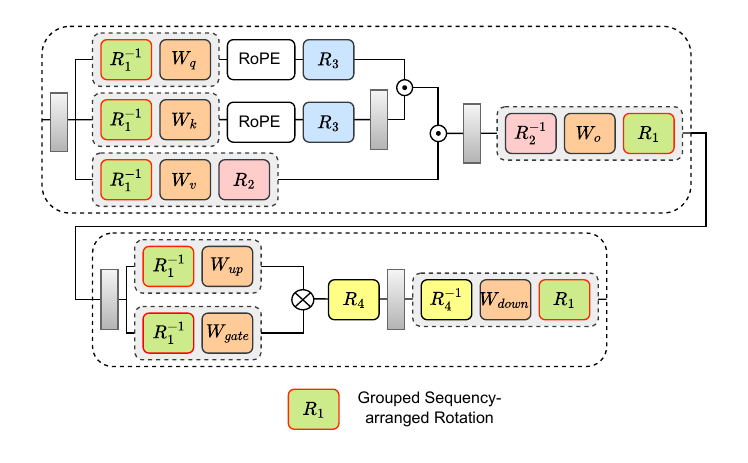} 
    \caption{Overall diagram of rotation scheme. We applied Grouped Sequency-arranged Rotation (GSR) on $R_1$. }
    \vspace{-0.5cm}
    \label{fig:main}
\end{figure}
Since a Hadamard matrix can be used as a rotation matrix when scaled and has an efficient algorithm, recent state-of-the-art methods make extensive use of the Hadamard transform~\cite{quarot,duarot,duquant,spinquant,ostquant}. We followed SpinQuant's terminology to describe our rotation scheme as Fig. \ref{fig:main}. At Fig. \ref{fig:main},  $R_1$ rotates all hidden activations between transformer blocks, $R_2$ rotates the value activation, $R_3$ rotates the query and key activations after RoPE, and $R_4$ rotates the input activation of the down projection. Specifically for $R_1$, a Randomized Hadamard Transform (RHT) is employed following the proposition in Quip\#~\cite{quipsharp} for better incoherence processing. This way, the outliers in the activation distribution are largely suppressed, achieving deployable W4A4KV4\footnote{We notate x-bit weight, y-bit activation, z-bit KV-cache into WxAyKVz like W4A4KV4.} performance on famous LLM models.

\section{Methodology}

\subsection{Grouped Sequency-arranged Rotation}
We propose Grouped Sequency-arranged Rotation (GSR), a training-free rotation technique to improve post-training quantization of LLMs under extreme quantization settings such as W2 and W2A4\footnote{Since 2-bit per-channel quantization can easily fail to converge, we assume group quantization in all cases.}. We denote the input and output channels of a weight $W\in\mathbb{R}^{C\times H}$ with $C$ and $H$. $G$ and $N$ denote the group size and the number of groups, respectively, so that $C=NG$.

As exhibited in Fig. \ref{fig:main}, we design a signal processing-inspired rotation matrix that can independently be plugged into existing rotation-based PTQ algorithms, as follows:
\begin{equation}\label{eqn:gsr}
R_{GSR} = \begin{bmatrix}
H_{wal}     & \textbf{0}    & \cdots    & \cdots    & \textbf{0}   \\
\textbf{0}  & H_{wal}       & \textbf{0}    & \cdots    & \vdots             \\
\vdots  & \textbf{0}    & \ddots        & \textbf{0}         & \vdots       \\
\vdots      & \vdots        & \textbf{0}        & \ddots    & \textbf{0}   \\
\textbf{0}  & \cdots        & \cdots        & \textbf{0}& H_{wal}
\end{bmatrix}
\end{equation},
where $H_{wal}\in\{-1, 1\}^{G \times G}$ is a ${G \times G}$ Walsh matrix, with $G$ being the quantization group size, and $\textbf{0}$ is a $G \times G$ zero matrix.

The proposed $R_{GSR}$ has several advantages over the RHT and the SpinQuant matrices:
First, like QuaRot~\cite{quarot}, it can replace any rotation matrix in existing PTQ methods without training for virtually free, as the only additional operation required is to pre-process a Sylvester-constructed Hadamard matrix to a Walsh matrix and apply the Kronecker product with an identity matrix before going into quantization. Second, it can systematically reduce weight quantization error by strategically arranging sequency filters with similar yet diverse sequency values (Section \ref{sec:sequency_analysis}). Third, it can also serve as an enhanced initialization for training-based methods such as SpinQuant~\cite{spinquant} and OSTQuant~\cite{ostquant} (Section \ref{sec:main_results}).

\subsection{The Effect of Sequency Arrangement on Group Quantization}\label{sec:sequency_analysis}
To justify our design, we investigate how the sequency ordering in our GSR can improve group quantization on weights. As shown in Fig. \ref{fig:main}, the weights are rotated twice as follows:
\begin{equation}\label{eqn:rweightdef}
W' = R_f^{-1}WR_r,
\end{equation}
where $R_f$ and $R_r$ are rotation matrices applied to the front and rear side of a weight $W$, respectively. For query weight $W_q$ as an example, $R_f=R_1$ and $R_r=I$ hold. We do not consider local rotation in this section for brevity.

An $(i, j)$ element of the rotated weight ($W'[i, j]$) can be expressed as follows:
\begin{equation}\label{eqn:rweightdetail}
\begin{aligned}
W'[i,j] &= \langle (R_f^{-1}W)[i,:], R_r[:, j] \rangle \\
= &\left\langle \left[ \langle R_f^{-1}[i, :], W[:, 1]\rangle, \langle R_f^{-1}[i, :], W[:, 2]\rangle, \right. \right. \\
&\left. \left. \dots, \langle R_f^{-1}[i, :], W[:, H]\rangle \right], R_r[:, j] \right\rangle.
\end{aligned}
\end{equation}
An $n$-th row group in $W'$ can be expressed as $W'[nG:(n+1)G,:]$, which leads to our observation \#1 by simply substituting $i$ to $nG:(n+1)G$ in Eqn. \ref{eqn:rweightdetail}.
\begin{observation}{}
Under group quantization, each column group in the front rotation matrix $R_f$ generates distinct rotated weight groups, and all columns in the rear rotation matrix $R_r$ are always applied to all rows in the original weight.

In other words, a group in the rotated weight $W'$ is the original weight transformed by the corresponding group of filters in the front rotation matrix and then by all filters in the rear rotation matrix.
\end{observation}

\paragraph{Comparing Hadamard and Walsh} Now, we relate the sequency arrangement to group quantization performance. For $R_r$, the arrangement has no impact as long as the set of sequency values is equal, which is the case with comparing the Hadamard and Walsh matrices. Therefore, we focus on $R_f$. The Walsh matrix (with the sequency ordering) has smaller sequency variance within each column group than the Hadamard matrix because the sequency values increase linearly. Since sequency is analogous to frequency in the conventional frequency-domain filtering, the Walsh matrix will produce rotated weight groups with fewer massive outliers. As shown in Table \ref{tab:rotation_config}, $R_1$ works as $R_f$ on many different types of transformer weights including $W_q,W_k,W_v,W_{up},\ \text{and}\ W_{gate}$,   changing $R_1$ from Hadamard to Walsh helps reduce the quantization error for these weights.

\begin{table}[]
\centering
\resizebox{\columnwidth}{!}{%
\begin{tabular}{@{}cccccccc@{}}
\toprule
Weight & $W_q$ & $W_k$ & $W_v$ & $W_o$ & $W_{up}$ & $W_{gate}$ & $W_{down}$ \\ \midrule
$R_f$  & $R_1$ & $R_1$ & $R_1$ & $R_2$ & $R_1$    & $R_1$      & $R_4$      \\
$R_r$  & $I$   & $I$   & $R_2$ & $R_1$ & $I$      & $I$        & $R_1$      \\ \bottomrule
\end{tabular}%
}
\caption{Rotation matrix configuration on each weight type in LLaMA-like transformer architecture. $I$ is the identity matrix.}
\label{tab:rotation_config}
\end{table}

\paragraph{Comparing RHT and Walsh}
The randomization method in Quip\#~\cite{quipsharp} and QuaRot~\cite{quarot} only flips the signs of diagonal elements in a Hadamard matrix. This process keeps the overall sequency arrangement with no significant changes. Therefore, we can compare the RHT against the Walsh following the same logic as in the previous section.

\subsection{Global vs. Local Rotation}
\begin{figure}[h!]
    \centering

    \begin{subfigure}[t]{\linewidth} 
        \centering
        \includegraphics[width=1.0\linewidth]{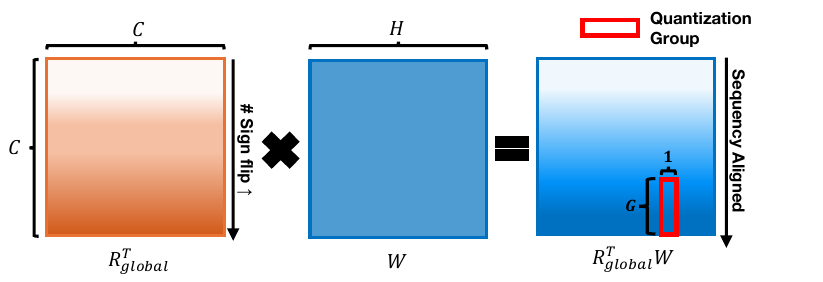} 
        \caption{Global rotation applies a full-matrix transformation across all dimensions and spreads outlier effects widely.}
        \label{fig:global_rot}
    \end{subfigure}    
    \vspace{1em} 
    
    \begin{subfigure}[t]{\linewidth} 
        \centering
        \includegraphics[width=1.0\linewidth]{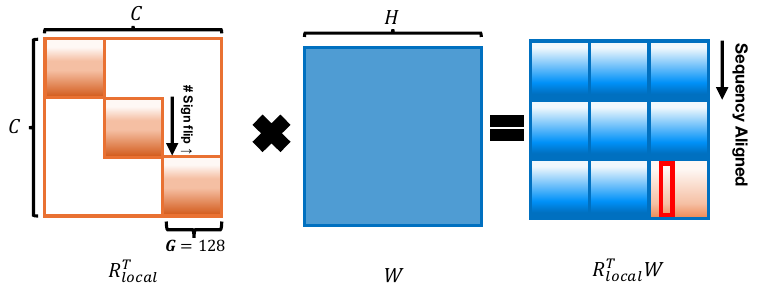} 
        \caption{Local rotation applies block-diagonal transformations within groups and confines outlier effects within each block. For illustration purposes, three blocks are depicted, while the actual number of blocks is given by $N=C/G$.}
        \label{fig:local_rot}
    \end{subfigure}
    \caption{Overview of global and local rotation strategies. Global rotation transforms the entire space and amplifies outlier effects and local rotation advances control over outliers within blocks to improve quantization robustness.}
    \vspace{-0.3cm}
    \label{fig:global_vs_local}
\end{figure}

Local rotation (using block-diagonal matrices) is generally more effective than global rotation (using a single large matrix)~\cite{duquant,duarot,dfrot}. Global rotation can struggle to effectively handle outliers, whether in activations or weights, as it spreads their impact to the whole input channel.
 Local rotation, however, confines the effects of such outliers within their specific block or group as in Fig. \ref{fig:global_vs_local} (b). When used with the Walsh matrix, this containment helps better reduce errors, which is also beneficial for low-bit weight quantization.

\begin{table*}[!ht]
\centering
\setlength{\tabcolsep}{5pt}
\renewcommand{\arraystretch}{1.3}
\resizebox{1.0\textwidth}{!}{%
{\normalsize
\begin{tabular}{ccccc|ccccc|ccccc}

\multicolumn{5}{c}{} & \multicolumn{5}{c}{} & \multicolumn{5}{c}{} \\
\toprule[2pt]
\textbf{Method} & \textbf{Bits} & \textbf{$R_1$} & \textbf{PPL$^{\downarrow}$} &\textbf{0-shot$^{\uparrow}$} & \textbf{Method} & \textbf{Bits} & \textbf{$R_1$} & \textbf{PPL$^{\downarrow}$} & \textbf{0-shot$^{\uparrow}$} & \textbf{Method} & \textbf{Bits} & \textbf{$R_1$} & \textbf{PPL$^{\downarrow}$} & \textbf{0-shot$^{\uparrow}$} \\
\midrule
& W16A16 & & 5.47 & 69.81 &  & W16A16 & & 5.47 & 69.81 & & W16A16 & & 5.47 & 65.21  \\
\midrule
QuaRot & W2A16 & GH & 20.29 & 32.06 & SpinQuant & W2A16 & GH & 16.45 & 31.04 & OSTQuant & W2A16 & GH & 10.97 & 45.52 \\
 &  & GW & \textbf{15.38} & \textbf{39.30} &  &  & GW & \textbf{16.44} & \textbf{34.52} &  &  & GW & \textbf{9.51} & \textbf{46.83} \\
 &  & LH & 12.11 & 41.01 &  &  & LH & 13.17 & 39.84 &  &  & LH & 9.16 & 49.84 \\
 &  & GSR & \textbf{11.59} & \textbf{42.44} &  &  & GSR & \textbf{12.04} & \textbf{42.11} &  &  & GSR & \textbf{9.03} & \textbf{50.51} \\ \midrule
 QuaRot& W2A4  & GH & 31.33 & 27.87 & SpinQuant & W2A4  & GH & 22.94 & 31.77 & OSTQuant & W2A4  & GH & 16.16 & 38.18 \\
 &   & GW & \textbf{20.34} & \textbf{33.75} &  &   & GW & \textbf{18.86} & \textbf{32.05} &  &   & GW & \textbf{14.68} & \textbf{40.67} \\
 &   & LH & 17.74 & 36.88 &  &   & LH & 15.79 & 34.57 &  &   & LH & 12.44 & 43.69 \\
 &   & GSR & \textbf{15.23} & \textbf{37.89} &  &   & GSR & \textbf{15.47} & \textbf{34.75} &  &   & GSR & \textbf{11.77} & \textbf{44.56} \\
\bottomrule[2pt]
\end{tabular}
}
}
\vspace{-0.3cm}
\caption{Comparison of the perplexity score on WikiText-2 and the averaged accuracy on zero-shot common-sense reasoning tasks. This experiment presents a comparative analysis across different methods to elucidate the performance differences arising from the types of rotation matrices employed. In the $R_1$ column, the notations "G", "L", and "H" correspond to global, local, and Hadamard, respectively. For example, 'GH' indicates that a global Hadamard rotation is applied to $R_1$.
}
\vspace{-0.5cm}
\label{tab:main}
\end{table*}

\section{Experimental Results}

\paragraph{Baseline}
We conducted experiments to assess whether the proposed GSR offers improved performance over previously used rotation matrices. Comparisons were made across QuaRot, SpinQuant, and OSTQuant. To ensure a fair evaluation, 
all methods were assessed by applying group quantization to their originally reported quantization configurations, under W2A16 and W2A4 settings. Changes in rotation, such as switching to the Walsh matrix or applying local rotation, were applied only to $R_1$, as further analyzed in the Appendix \ref{sec:appendix_ablation}. Details of the quantization configurations are provided in the Appendix \ref{sec:appendix_impl_details}.

\paragraph{Model and Datasets}
The proposed method was evaluated on Llama-2-7B~\cite{llama2}. To assess general language modeling capability, we measured PPL on WikiText-2~\cite{wiki2} with a context length of 2048 tokens. To evaluate reasoning ability, we conducted common zero-shot evaluations on a set of reasoning tasks, following the same datasets used in baseline methods. Specifically, QuaRot and SpinQuant were evaluated on Arc (Easy and Challenge)~\cite{arc}, HellaSwag~\cite{hellaswag}, LAMBADA~\cite{lambada}, PIQA~\cite{piqa}, and WinoGrande~\cite{winogrande}, while OSTQuant was additionally evaluated on BoolQ~\cite{boolq}, OpenBookQA~\cite{openbookqa}, and SIQA~\cite{siqa}.

\paragraph{Implementation Details and Overall Results}\label{sec:main_results}
We denote the global Hadamard matrix as GH, the global Walsh matrix as GW, local Hadamard matrix as LH. All Hadamard matrices are randomized, following common practice in previous rotation-based algorithms. When constructing Walsh matrices, the original Hadamard matrix is used. The other details not mentioned here are listed in the Appendix \ref{sec:appendix_impl_details}.

The overall results are summarized in Table~\ref{tab:main}. Across all methods, our proposed approach consistently outperforms the GH, achieving lower PPL and higher accuracy on reasoning tasks. In particular, applying the GW to QuaRot (i.e., re-ordering rows of the Hadamard matrix with natural ordering) yields approximately 1 point lower PPL compared to SpinQuant, validating the benefit of the sequency arrangement. Given that SpinQuant typically consumes much greater computational costs than QuaRot, this result suggests that adopting GSR enables QuaRot to achieve superior performance and efficiency. While OSTQuant learns both the rotation matrix and the smooth factor through optimization and achieves a PPL of 10.97 in the W2 setting, QuaRot with GSR attains a comparable PPL of 11.59 by simply replacing $R_1$in a training-free manner. In the W2A4 setting, QuaRot with GSR even surpasses OSTQuant, achieving a lower PPL of 15.23 compared to 16.16, indicating that better performance can be obtained with fewer resources. The effectiveness of GSR also holds when applied to OSTQuant, consistently leading to further performance gains.

The advantage of the sequency arrangement is enhanced when paired with the local rotation. When comparing the LH and GSR on QuaRot, GSR consistently also delivers better performance across all cases, similar to the improvements observed in global rotation (GH vs GW).
Moreover, in zero-shot task evaluations, the Walsh matirx consistently outperforms the Hadamard. Notably, in the QuaRot W2 setting, the GW achieves approximately 7 points higher accuracy compared to the GH, again surpassing SpinQuant.
Complete individual scores for each task are provided in Appendix~\ref{sec:appendix_full_results}.

\section{Conclusion}
In this paper, we proposed a novel training-free rotation technique, Grouped Sequency-arranged Rotation (GSR), inspired by signal processing theory on Walsh-Hadamard transform and sequency. The GSR makes use of the Walsh matrix to place transformed weights filtered by similar sequency values closer, and combines the local rotation idea for constraining possible remaining outliers within a single quantization group per row. A theoretical justification is also provided for each component. Experimental results verify the effectiveness of our proposed method on common benchmarks for LLM quantization, including WikiText-2 and popular zero-shot common-sense reasoning tasks.

\section*{Limitations}
Our proposed method has proven effective only under extremely low-bit weight quantization with group quantization. On larger bit configurations, the quantization error becomes much less significant, so that the sequency alignment cannot show visible improvement. In addition, to ensure the generalizability of our approach, we plan to extend our experiments to other model architectures and datasets in future work.

\section*{Acknowledgments}
This work was supported by Samsung Advanced Institute of Technology, and MX division, Samsung Electronics Co., Ltd., Inter-university Semiconductor Research Center (ISRC) and Institute of Information \& communications Technology Planning \& Evaluation (IITP) grant funded by the Korea government(MSIT) [NO. RS-2021-II211343, Artificial Intelligence Graduate School Program (Seoul National University)].

\bibliography{custom}

\begin{thebibliography}{19}
\providecommand{\natexlab}[1]{#1}

\bibitem[{Ashkboos et~al.(2024)Ashkboos, Mohtashami, Croci, Li, Jaggi, Alistarh, Hoefler, and Hensman}]{quarot}
Saleh Ashkboos, Amirkeivan Mohtashami, Maximilian~L Croci, Bo~Li, Martin Jaggi, Dan Alistarh, Torsten Hoefler, and James Hensman. 2024.
\newblock \href {https://arxiv.org/abs/2405.04517} {Quarot: Outlier-free 4-bit inference in rotated llms}.
\newblock In \emph{Thirty-eighth Conference on Neural Information Processing Systems}.

\bibitem[{Bisk et~al.(2020)Bisk, Zellers, Gao, Choi et~al.}]{piqa}
Yonatan Bisk, Rowan Zellers, Jianfeng Gao, Yejin Choi, and 1 others. 2020.
\newblock Piqa: Reasoning about physical commonsense in natural language.
\newblock In \emph{Proceedings of the AAAI conference on artificial intelligence}, volume~34, pages 7432--7439.

\bibitem[{Clark et~al.(2019)Clark, Lee, Chang, Kwiatkowski, Collins, and Toutanova}]{boolq}
Christopher Clark, Kenton Lee, Ming-Wei Chang, Tom Kwiatkowski, Michael Collins, and Kristina Toutanova. 2019.
\newblock \href {https://doi.org/10.18653/v1/N19-1300} {{B}ool{Q}: Exploring the surprising difficulty of natural yes/no questions}.
\newblock In \emph{Proceedings of the 2019 Conference of the North {A}merican Chapter of the Association for Computational Linguistics: Human Language Technologies, Volume 1 (Long and Short Papers)}, pages 2924--2936, Minneapolis, Minnesota. Association for Computational Linguistics.

\bibitem[{Clark et~al.(2018)Clark, Cowhey, Etzioni, Khot, Sabharwal, Schoenick, and Tafjord}]{arc}
Peter Clark, Isaac Cowhey, Oren Etzioni, Tushar Khot, Ashish Sabharwal, Carissa Schoenick, and Oyvind Tafjord. 2018.
\newblock Think you have solved question answering? try arc, the ai2 reasoning challenge.
\newblock \emph{arXiv:1803.05457v1}.

\bibitem[{Frantar et~al.(2022)Frantar, Ashkboos, Hoefler, and Alistarh}]{gptq}
Elias Frantar, Saleh Ashkboos, Torsten Hoefler, and Dan Alistarh. 2022.
\newblock Optq: Accurate quantization for generative pre-trained transformers.
\newblock In \emph{The Eleventh International Conference on Learning Representations}.

\bibitem[{Hu et~al.(2025)Hu, Cheng, Yang, Chen, Xu, JiangyongYu, XUCHEN, Yuan, jiang, and Zhou}]{ostquant}
Xing Hu, Yuan Cheng, Dawei Yang, Zhixuan Chen, Zukang Xu, JiangyongYu, XUCHEN, Zhihang Yuan, Zhe jiang, and Sifan Zhou. 2025.
\newblock \href {https://openreview.net/forum?id=rAcgDBdKnP} {{OSTQ}uant: Refining large language model quantization with orthogonal and scaling transformations for better distribution fitting}.
\newblock In \emph{The Thirteenth International Conference on Learning Representations}.

\bibitem[{Lin et~al.(2024)Lin, Xu, Wu, Cui, Zhang, Mou, Song, Sun, and Wei}]{duquant}
Haokun Lin, Haobo Xu, Yichen Wu, Jingzhi Cui, Yingtao Zhang, Linzhan Mou, Linqi Song, Zhenan Sun, and Ying Wei. 2024.
\newblock Duquant: Distributing outliers via dual transformation makes stronger quantized llms.
\newblock \emph{Advances in Neural Information Processing Systems}, 37:87766--87800.

\bibitem[{Liu et~al.(2025)Liu, Zhao, Fedorov, Soran, Choudhary, Krishnamoorthi, Chandra, Tian, and Blankevoort}]{spinquant}
Zechun Liu, Changsheng Zhao, Igor Fedorov, Bilge Soran, Dhruv Choudhary, Raghuraman Krishnamoorthi, Vikas Chandra, Yuandong Tian, and Tijmen Blankevoort. 2025.
\newblock \href {https://openreview.net/forum?id=ogO6DGE6FZ} {Spinquant: Llm quantization with learned rotations}.
\newblock In \emph{The Thirteenth International Conference on Learning Representations}.

\bibitem[{Merity et~al.(2017)Merity, Xiong, Bradbury, and Socher}]{wiki2}
Stephen Merity, Caiming Xiong, James Bradbury, and Richard Socher. 2017.
\newblock \href {https://openreview.net/forum?id=Byj72udxe} {Pointer sentinel mixture models}.
\newblock In \emph{International Conference on Learning Representations}.

\bibitem[{Mihaylov et~al.(2018)Mihaylov, Clark, Khot, and Sabharwal}]{openbookqa}
Todor Mihaylov, Peter Clark, Tushar Khot, and Ashish Sabharwal. 2018.
\newblock \href {https://doi.org/10.18653/v1/D18-1260} {Can a suit of armor conduct electricity? a new dataset for open book question answering}.
\newblock In \emph{Proceedings of the 2018 Conference on Empirical Methods in Natural Language Processing}, pages 2381--2391, Brussels, Belgium. Association for Computational Linguistics.

\bibitem[{Paperno et~al.(2016)Paperno, Kruszewski, Lazaridou, Pham, Bernardi, Pezzelle, Baroni, Boleda, and Fern{\'a}ndez}]{lambada}
Denis Paperno, Germ{\'a}n Kruszewski, Angeliki Lazaridou, Ngoc~Quan Pham, Raffaella Bernardi, Sandro Pezzelle, Marco Baroni, Gemma Boleda, and Raquel Fern{\'a}ndez. 2016.
\newblock \href {https://doi.org/10.18653/v1/P16-1144} {The {LAMBADA} dataset: Word prediction requiring a broad discourse context}.
\newblock In \emph{Proceedings of the 54th Annual Meeting of the Association for Computational Linguistics (Volume 1: Long Papers)}, pages 1525--1534, Berlin, Germany. Association for Computational Linguistics.

\bibitem[{Sakaguchi et~al.(2021)Sakaguchi, Bras, Bhagavatula, and Choi}]{winogrande}
Keisuke Sakaguchi, Ronan~Le Bras, Chandra Bhagavatula, and Yejin Choi. 2021.
\newblock Winogrande: An adversarial winograd schema challenge at scale.
\newblock \emph{Communications of the ACM}, 64(9):99--106.

\bibitem[{Sap et~al.(2019)Sap, Rashkin, Chen, Le~Bras, and Choi}]{siqa}
Maarten Sap, Hannah Rashkin, Derek Chen, Ronan Le~Bras, and Yejin Choi. 2019.
\newblock \href {https://doi.org/10.18653/v1/D19-1454} {Social {IQ}a: Commonsense reasoning about social interactions}.
\newblock In \emph{Proceedings of the 2019 Conference on Empirical Methods in Natural Language Processing and the 9th International Joint Conference on Natural Language Processing (EMNLP-IJCNLP)}, pages 4463--4473, Hong Kong, China. Association for Computational Linguistics.

\bibitem[{Tam and Goulet(1972)}]{walsh}
Le~Dinh~Chon Tam and R.Y. Goulet. 1972.
\newblock \href {https://doi.org/10.1109/T-C.1972.223524} {On arithmetical shift for walsh functions}.
\newblock \emph{IEEE Transactions on Computers}, C-21(12):1451--1452.

\bibitem[{Touvron et~al.(2023)Touvron, Martin, Stone, Albert, Almahairi, Babaei, Bashlykov, Batra, Bhargava, Bhosale et~al.}]{llama2}
Hugo Touvron, Louis Martin, Kevin Stone, Peter Albert, Amjad Almahairi, Yasmine Babaei, Nikolay Bashlykov, Soumya Batra, Prajjwal Bhargava, Shruti Bhosale, and 1 others. 2023.
\newblock Llama 2: Open foundation and fine-tuned chat models.
\newblock \emph{arXiv preprint arXiv:2307.09288}.

\bibitem[{Tseng et~al.(2024)Tseng, Chee, Sun, Kuleshov, and Sa}]{quipsharp}
Albert Tseng, Jerry Chee, Qingyao Sun, Volodymyr Kuleshov, and Christopher~De Sa. 2024.
\newblock \href {https://openreview.net/forum?id=9BrydUVcoe} {Qu{IP}\${\textbackslash}\#\$: Even better {LLM} quantization with hadamard incoherence and lattice codebooks}.
\newblock In \emph{Forty-first International Conference on Machine Learning}.

\bibitem[{Xiang and Zhang(2024)}]{dfrot}
Jingyang Xiang and Sai~Qian Zhang. 2024.
\newblock Dfrot: Achieving outlier-free and massive activation-free for rotated llms with refined rotation.
\newblock \emph{arXiv preprint arXiv:2412.00648}.

\bibitem[{Xiang et~al.(2025)Xiang, Zhang, Ma, Wang, yulei, LiuChuan, Lin, and Liu}]{duarot}
Jingyang Xiang, Ying Zhang, Chi Ma, Yujie Wang, yulei, LiuChuan, Wei Lin, and Yong Liu. 2025.
\newblock \href {https://openreview.net/forum?id=oHBS7R6JcP} {Duarot: Dual rotation for advanced outlier mitigation in rotated {LLM}s}.

\bibitem[{Zellers et~al.(2019)Zellers, Holtzman, Bisk, Farhadi, and Choi}]{hellaswag}
Rowan Zellers, Ari Holtzman, Yonatan Bisk, Ali Farhadi, and Yejin Choi. 2019.
\newblock \href {https://doi.org/10.18653/v1/P19-1472} {{H}ella{S}wag: Can a machine really finish your sentence?}
\newblock In \emph{Proceedings of the 57th Annual Meeting of the Association for Computational Linguistics}, pages 4791--4800, Florence, Italy. Association for Computational Linguistics.

\end{thebibliography}

\appendix

\section{Appendix}
\label{sec:appendix}
\subsection{Additional Implementation Details}
\label{sec:appendix_impl_details}
For a fair comparison, only group quantization was additionally applied, while the primary quantization settings originally reported for each method were preserved. The detailed settings applied to each method are described below.
\paragraph{GPTQ}
During weight quantization with GPTQ~\cite{gptq}, the calibration was performed by sampling 128 contexts, each consisting of 2048 tokens, from the WikiText2 dataset.

\paragraph{QuaRot}
For QuaRot~\cite{quarot}, GPTQ-based quantization was applied with asymmetric weight quantization, MSE-based clipping, and group quantization using a group size of 128. Activation quantization was performed using symmetric round-to-nearest (RTN) quantization with a clipping ratio of 0.9 and a group size of 128.

\paragraph{SpinQuant}
For SpinQuant~\cite{spinquant}, since GPTQ was used during PTQ, weight quantization was not applied during the rotation matrix training phase. However, when activation quantization was included, activation quantization-aware training was performed using an RTN quantizer, with symmetric quantization and a group size of 128 applied to activations.

\paragraph{OSTQuant}
For OSTQuant~\cite{ostquant}, both the rotation matrix and the smoothing factor were learned. During weight-only quantization, weight-quantization-aware training was conducted using asymmetric quantization, MSE-based clipping, and a group size of 128. When quantizing both weights and activations, the weights were kept frozen, and only the effect of activation RTN quantization was considered, with a group size of 128 applied.

\subsection{Ablation Study}
\label{sec:appendix_ablation}
\paragraph{Global and Local Rotation on $R_4$}
\begin{table}[htbp]
\centering
{\small
\begin{tabular}{c|c|c|c|c}
\hline
\textbf{Method} & \textbf{$R_1$} & \textbf{$R_4$} & \textbf{PPL} & \textbf{PPL$^\dagger$} \\ \hline
\multirow{4}{*}{QuaRot} & LH & GH & 12.11 & 17.74 \\ 
                       & LH & LH & 12.65 & 14.64 \\ 
                       & GSR & GH & 11.59 & 15.23 \\ 
                       & GSR & LH & 11.22 & 13.83 \\ \hline
\end{tabular}
\caption{Ablation results on the effect of local rotation for $R_4$ in Llama-2-7B. PPL represents the results for W2, and PPL$^\dagger$ represents the results for W2A4.}
\label{tab:abl}
}
\end{table}

As part of the ablation study, we applied local rotation to $R_4$, originally using global rotation. Table~\ref{tab:abl} shows that local rotation consistently improves performance under activation quantization (W2A4), but has negligible impact under weight-only quantization (W2).

Given the role and placement of $R_4$, it primarily rotates activation outliers through an online rotation mechanism before input activations enter the down-projection of the FFN layer. From the weight perspective, since $R_1$ and $R_4$ are fused into the weights during inference, the benefit of local rotation is realized only once. Thus, the performance gains observed from modifications to $R_4$ can be attributed mainly to the activation quantization process.

Nonetheless, applying local rotation to the online rotation introduces practical challenges. In particular, it disables the use of the fast-hadamard-transform, requiring the entire FP32 matrix tensor to be stored in memory during inference, which is impractical. We left addressing this limitation for future work.

\subsection{Complete Reasoning Tasks Results}
\label{sec:appendix_full_results}
In this section, Table~\ref{tab:complete_first} and Table~\ref{tab:complete_second} present evaluation results for each zero-shot task.
\begin{table*}[htbp]
    \centering
    \setlength{\tabcolsep}{5pt}
    \renewcommand{\arraystretch}{1.0}
    \resizebox{\textwidth}{!}{%
    {\normalsize
    \begin{tabular}{ccc|ccccccccccc}
        \toprule[1pt]
         \raisebox{-4ex}{ \textbf{\#Bits}} & \multicolumn{2}{c}{ \raisebox{-2ex}{\textbf{Configuration}} } & \multicolumn{1}{c}{ \raisebox{-5ex}{\textbf{ARC-c}}} & \multicolumn{1}{c}{ \raisebox{-5ex}{\textbf{ARC-e}}} & \multicolumn{1}{c}{ \raisebox{-5ex}{\textbf{Hella.}}} & \multicolumn{1}{c}{ \raisebox{-5ex}{\textbf{lambada}}} & \multicolumn{1}{c}{ \raisebox{-5ex}{\textbf{lambada-o}}} & \multicolumn{1}{c}{ \raisebox{-5ex}{\textbf{lambada-s}}} & \multicolumn{1}{c}{ \raisebox{-5ex}{\textbf{PIQA}}}& \multicolumn{1}{c}{ \raisebox{-5ex}{\textbf{Wino.}}} & \multicolumn{1}{c}{ \raisebox{-5ex}{\textbf{Avg.}}} \\
        \cmidrule(lr){2-3} 
         & \textbf{Method} & \textbf{$R_1$} & \\
        \midrule
         \multirow{1}{*}{16-16} &  &  & 46.25 & 74.58 & 75.99 & 71.12 & 73.92 & 68.33 & 79.11 & 69.14 & 69.81   \\
        \cmidrule(lr){1-12}
         \multirow{4}{*}{2-16} 
         & \multirow{4}{*}{QuaRot} & GH & 23.04 & 43.27 & 35.51 & 13.33 & 14.48 & 12.19 & 59.14 & 55.49 & 32.06   \\
         & & GW &  25.94 & 44.49 & 42.07 & 27.88 & 30.53 & 25.23 & 61.26 & 56.99 & 39.30    \\
         & & LH &  27.22 & 48.91 & 46.12 & 27.56 & 30.18 & 24.94 & 66.38 & 56.75 & 41.01  \\
         & & GSR &  26.79 & 49.71 & 47.86 & 30.90 & 35.46 & 26.35 & 64.85 & 57.62 & 42.44  \\
        \cmidrule(lr){1-12}
         \multirow{4}{*}{2-4} 
         &\multirow{4}{*}{QuaRot} & GH & 21.67 & 35.31 & 33.00 & 8.64 & 9.72 & 7.55 & 57.13 & 49.96 & 27.87   \\
         & & GW & 22.78 & 38.34 & 36.56 & 19.75 & 22.49 & 17.00 & 58.81 & 54.30 & 33.75   \\
         & & LH &  25.77 & 43.94 & 41.20 & 22.52 & 23.95 & 21.09 & 62.62 & 53.91 & 36.88 \\
         & & GSR &  27.22 & 45.20 & 43.46 & 23.83 & 26.92 & 20.75 & 61.64 & 54.14 & 37.89 \\
        \cmidrule(lr){1-12}
         \multirow{4}{*}{2-16} 
         & \multirow{4}{*}{SpinQuant} & GH & 22.70 & 41.29 & 34.37 & 12.65 & 14.26 & 11.04 & 57.83 & 54.14 & 31.04   \\
         & & GW & 22.70 & 40.82 & 36.57 & 20.98 & 21.41 & 20.55 & 59.19 & 53.91 & 34.52    \\
         & & LH &  25.43 & 45.58 & 42.43 & 28.58 & 31.34 & 25.81 & 63.17 & 56.35 & 39.84 \\
         & & GSR &  25.34 & 46.46 & 44.90 & 32.73 & 34.95 & 30.51 & 64.31 & 57.70 & 42.11 \\
        \cmidrule(lr){1-12}
         \multirow{4}{*}{2-4} 
         & \multirow{4}{*}{SpinQuant} & GH & 24.23 & 38.97 & 34.68 & 14.36 & 15.74 & 12.98 & 57.13 & 56.04 & 31.77  \\
         & & GW & 22.78 & 37.04 & 33.75 & 17.70 & 20.32 & 15.08 & 57.13 & 52.57 & 32.05   \\
         & & LH &  23.89 & 40.28 & 39.80 & 19.25 & 21.08 & 17.43 & 60.61 & 54.22 & 34.57  \\
         & & GSR &  25.17 & 41.58 & 36.54 & 20.68 & 23.21 & 18.14 & 59.74 & 52.96 & 34.75  \\
        \bottomrule[1pt]
    \end{tabular}
    }
    }
    \caption{Complete comparison of accuracy on Zero-shot Common Sense Reasoning tasks for Llama2-7B with QuaRot and SpinQuant. \textbf{lambada-o} and \textbf{lambada-s} represent \textbf{lambada-openai} and \textbf{lambada-standard}, respectively.
}
\label{tab:complete_first}
\end{table*}

\begin{table*}[htbp]
    \centering
    \setlength{\tabcolsep}{5pt}
    \renewcommand{\arraystretch}{1.0}
    \resizebox{\textwidth}{!}{%
    {\normalsize
    \begin{tabular}{ccc|cccccccccccc}
        \toprule[1pt]
         \raisebox{-4ex}{ \textbf{\#Bits}} & \multicolumn{2}{c}{ \raisebox{-2ex}{\textbf{Configuration}} } & \multicolumn{1}{c}{ \raisebox{-5ex}{\textbf{ARC-c}}} & \multicolumn{1}{c}{ \raisebox{-5ex}{\textbf{ARC-e}}} & \multicolumn{1}{c}{ \raisebox{-5ex}{\textbf{boolq}}} & \multicolumn{1}{c}{ \raisebox{-5ex}{\textbf{Hella.}}} & \multicolumn{1}{c}{ \raisebox{-5ex}{\textbf{lambada-o}}} & \multicolumn{1}{c}{ \raisebox{-5ex}{\textbf{openbook-qa}}} & \multicolumn{1}{c}{ \raisebox{-5ex}{\textbf{PIQA}}}& \multicolumn{1}{c}{ \raisebox{-5ex}{\textbf{Social-IQA}}} & \multicolumn{1}{c}{ \raisebox{-5ex}{\textbf{Wino.}}} & \multicolumn{1}{c}{ \raisebox{-5ex}{\textbf{Avg.}}} \\
        \cmidrule(lr){2-3} 
         & \textbf{Method} & \textbf{$R_1$} & \\
        \midrule
         \multirow{1}{*}{16-16} &  &  & 46.42 & 74.33 & 77.71 & 75.94 & 73.69 & 44.20 & 79.16 & 45.91 & 69.53 & 65.21   \\
        \cmidrule(lr){1-13}
         \multirow{4}{*}{2-16} 
         & \multirow{4}{*}{OSTQuant} & GH & 23.63 & 50.38 & 62.87 & 34.75 & 40.19 & 19.60 & 63.44 & 36.85 & 59.04 & 45.52   \\
         & & GW & 25.00 & 53.79 & 63.15 & 36.16 & 39.14 & 19.80 & 65.61 & 38.33 & 59.43 & 46.83   \\
         & & LH & 27.56 & 57.53 & 63.30 & 39.47 & 50.96 & 20.00 & 66.76 & 39.36 & 59.98 & 49.84  \\
         & & GSR & 26.62 & 60.56 & 65.29 & 38.69 & 56.20 & 22.40 & 66.54 & 38.08 & 61.09 & 50.51  \\
        \cmidrule(lr){1-13}
         \multirow{4}{*}{2-4} 
         & \multirow{4}{*}{OSTQuant} & GH & 19.37 & 39.14 & 50.98 & 31.48 & 18.38 & 15.20 & 60.39 & 36.08 & 53.28 & 38.18   \\
         & & GW & 19.88 & 45.08 & 61.83 & 32.00 & 22.61 & 15.00 & 60.23 & 36.34 & 52.09 & 40.67   \\
         & & LH & 24.66 & 50.25 & 63.21 & 34.82 & 26.61 & 18.60 & 63.93 & 36.80 & 55.33 & 43.69 \\
         & & GSR & 23.21 & 51.89 & 62.81 & 35.05 & 33.75 & 18.40 & 63.28 & 37.72 & 56.59 & 44.56 \\
        \bottomrule[1pt]
    \end{tabular}
    }
    }
    \caption{Complete comparison of accuracy on Zero-shot Common Sense Reasoning tasks for Llama2-7B with OSTQuant. \textbf{lambada-o} represents \textbf{lambada-openai}.}
\label{tab:complete_second}
\end{table*}

\end{document}